\documentclass[letterpaper, 10 pt, conference]{ieeeconf}  
\usepackage[table,xcdraw]{xcolor}
\usepackage{graphicx}
\usepackage{array, booktabs, multirow}
\usepackage{amssymb}
\usepackage{pifont}
\usepackage{adjustbox}
\usepackage{amsmath}
\usepackage{tabularx}
\usepackage{comment}
\usepackage{hyperref}       
\usepackage{url}   

\IEEEoverridecommandlockouts                              

\title{\LARGE \bf
IndraEye: Infrared Electro-Optical UAV-based Perception Dataset for Robust Downstream Tasks
}

\author{Manjunath D$^{1}$, Prajwal Gurunath$^{2}$, Sumanth Udupa$^{2}$ , 
Aditya Gandhamal$^{3}$, \\ Shrikar Madhu$^{3}$,  Aniruddh Sikdar$^{4}$, Suresh Sundaram $^{5} $
\thanks{$^{1}$Manjunath D with the Department of Aerospace Engineering, Indian Institute of Science, Bangalore.
        {\tt\small manjunathd1@iisc.ac.in}}%
\thanks{$^{2}$Prajwal Gurunath and Sumanth Udupa are with the Department of Aerospace Engineering, Indian Institute of Science, Bangalore.
        {\tt\small prajwalg@iisc.ac.in}}%
\thanks{$^{3}$Aditya Gandhamal and Shrikar Madhu are with the Department of Aerospace Engineering, Indian Institute of Science, Bangalore.
        {\tt\small adityam1@iisc.ac.in, shrikarm@iisc.ac.in}}%
\thanks{$^{4}$Aniruddh Sikdar is with the Robert Bosch Centre for Cyber-Physical Systems, Indian Institute of Science, Bangalore.
        {\tt\small aniruddhss@iisc.ac.in}}%
\thanks{$^{5}$Suresh Sundaram is with the Department of Aerospace Engineering, Indian Institute of Science, Bangalore.
        {\tt\small vssuresh@iisc.ac.in}}%
}

\begin{document}

\maketitle
\thispagestyle{empty}
\pagestyle{empty}

\begin{abstract}

Deep neural networks (DNNs) have shown exceptional performance when trained on well-illuminated images captured by Electro-Optical (EO) cameras, which provide rich texture details. However, in critical applications like aerial perception, it is essential for DNNs to maintain consistent reliability across all conditions, including low-light scenarios where EO cameras often struggle to capture sufficient detail. Additionally, UAV-based aerial object detection faces significant challenges due to scale variability from varying altitudes and slant angles, adding another layer of complexity. Existing methods typically address only illumination changes or style variations as domain shifts, but in aerial perception, correlation shifts also impact DNN performance. In this paper, we introduce the IndraEye dataset, a multi-sensor (EO-IR) dataset designed for various tasks. It includes 5,612 images with 145,666 instances, encompassing multiple viewing angles, altitudes, seven backgrounds, and different times of the day across the Indian subcontinent. The dataset opens up several research opportunities, such as multimodal learning, domain adaptation for object detection and segmentation, and exploration of sensor-specific strengths and weaknesses. IndraEye aims to advance the field by supporting the development of more robust and accurate aerial perception systems, particularly in challenging conditions. IndraEye dataset is benchmarked with  object detection and semantic segmentation tasks. Dataset and source codes are available at \href{https://bit.ly/indraeye}{https://bit.ly/indraeye}.


\end{abstract}

\section{INTRODUCTION}


Aerial perceptual robustness is essential for Unmanned Aerial Vehicles (UAVs) to operate effectively in harsh and low-light conditions, which is critical for robotic vision. Advances in affordable drone cameras have made UAV-based object detection feasible for applications such as aerial perception \cite{paramanandham2018infrared}, search and rescue \cite{bravo2019use}, traffic monitoring \cite{heintz2007images, kanistras2013survey}, and mapping \cite{samad2013potential}, particularly in environments not well-covered by standard datasets. Although most deep learning models are optimized for visible light cameras due to the abundance of Electro-Optical (EO) datasets, Infrared (IR) cameras offer superior performance in challenging conditions by penetrating dust and smoke. Their unique spectral capabilities enable effective operation in low-visibility and low-light scenarios. Despite these benefits, comprehensive IR datasets remain scarce compared to the extensive collection of EO datasets. To improve the robustness of aerial perception systems, there is a critical need for datasets that combine both electro-optical (EO) and infrared (IR) data for model training. To address this the proposed dataset contains (EO-IR pairs) images collected at various slant angles at different locations imaging both EO and IR simultaneously.

\begin{figure}[t]
    \centering
    \includegraphics[scale=0.188]{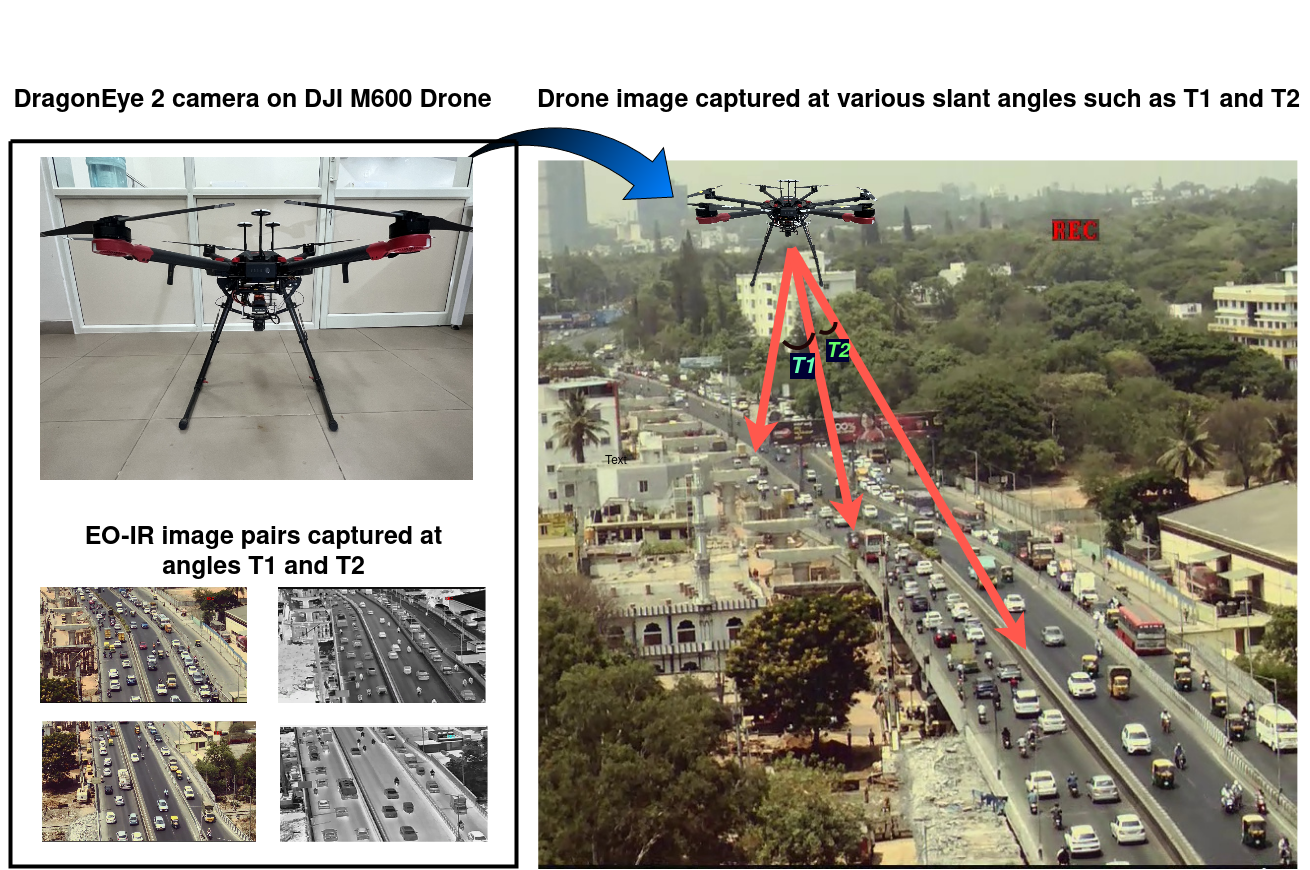}
    \caption{ Drone captured EO and IR image pairs at high altitude and look angles (T1 and T2) with varying slant perspectives, enhancing multi-modal aerial perception of semi-urban traffic scene.}
    \label{fig:inra_dronw}
\end{figure}

Many multi-modal (EO-IR) datasets, such as FLIR \cite{fliresh}, TarDAL \cite{liu2022target} and InfraParis \cite{franchi2024infraparis}, primarily focus on vehicle classes and road scenes, captured with cameras mounted on vehicles for autonomous driving applications. These datasets offer only low-altitude views. Although deep learning techniques have enhanced performance for these datasets \cite{sikdar2023deepmao}, high-altitude perspectives offer significant benefits for UAVs. Unlike fixed-viewpoint visual perception cameras, drone-based cameras can capture objects from various angles due to gimbal adjustments, which presents additional challenges for deep neural network (DNN) object detection algorithms \cite{saadiyean2024learning}. Objects viewed from directly above (nadir) appear differently compared to those observed from an off-nadir angle. Datasets such as DOTA \cite{xia2018dota} (EO) and Vedai \cite{razakarivony2016vehicle} (EO-IR pair) only provide nadir views. Deep learning models often struggle when trained on nadir views and tested on off-nadir angles \cite{weir2019spacenet}, making datasets like DOTA unsuitable for slant-angle aerial perception. The performance decline is even more pronounced for moderate to extreme off-nadir angles \cite{suo2023hit} \cite{weir2019spacenet} \cite{xia2018dota}. Additionally, off-nadir angles cause distant objects to appear smaller compared to those closer to the camera. In search and rescue operations, off-nadir views are utilized to cover larger areas, with objects appearing differently as the camera moves toward nadir. This scale variation becomes more pronounced with increased altitude, complicating UAV-based object detection. Thus, there is a need for datasets that capture a spectrum of nadir to off-nadir angles to enhance the robustness of neural networks against variations in angle and scale in aerial perception.

Drones equipped with stabilization gimbals, adjustable angles, and both EO and IR sensors are capable of capturing high-altitude data with broader fields of view. As multi-modal deep learning techniques advance \cite{liu2022target} \cite{zhao2023cddfuse}, access to EO-IR data across diverse scenes is increasingly critical for effective object recognition in complex environments. Existing datasets, such as CARPK \cite{hsieh2017drone}, UAV123 \cite{chen2022neighbortrack}, AU-AIR \cite{bozcan2020air}, and VisDrone \cite{cao2021visdrone}, are primarily focused on EO data, which often falls short for night-time aerial perception due to the lack of visible light. Therefore, extending beyond the visible spectrum is crucial for capturing comprehensive object details throughout the day and night. Conversely, while the HIT-UAV \cite{suo2023hit} dataset provides valuable IR data, it is unsuitable for daytime use where EO data is preferred for its richer detail. Additionally, HIT-UAV's focus limits the use of EO-to-IR domain adaptation \cite{akkaya2021self} \cite{cao2023contrastive} \cite{gan2023unsupervised} \cite{li2022cross} and sensor fusion techniques \cite{sikdar2024skd, zhao2023cddfuse}. Given these gaps, there is a clear need for a dataset that addresses slant-angle views, scale variations, day-night transitions, and multi-modal aspects to meet the complexities of aerial perception. Table 1 provides a comparison of the proposed IndraEye dataset’s features such as its range of illumination, class diversity, and background conditions against other leading contemporary aerial object perception datasets. Unlike other datasets, which focus solely on object detection, IndraEye also includes labels for both object detection and semantic segmentation along with rich textual prompts for VML techniques.

Foundation models have predominantly been trained on open-world data for image-level tasks \cite{cho2024cat}. To adapt these models for drone-based aerial perception tasks that involve multi-modal data, fine-tuning Vision-Language Models (VLMs) is crucial for enhancing their performance in object detection and segmentation. Research such as \cite{he2024prompting} demonstrates the benefits of integrating diverse data modalities. However, to fully leverage these advantages, access to comprehensive and well-annotated multi-modal datasets is necessary.

\begin{figure*}[t]
    \centering
    \includegraphics[scale=0.385]{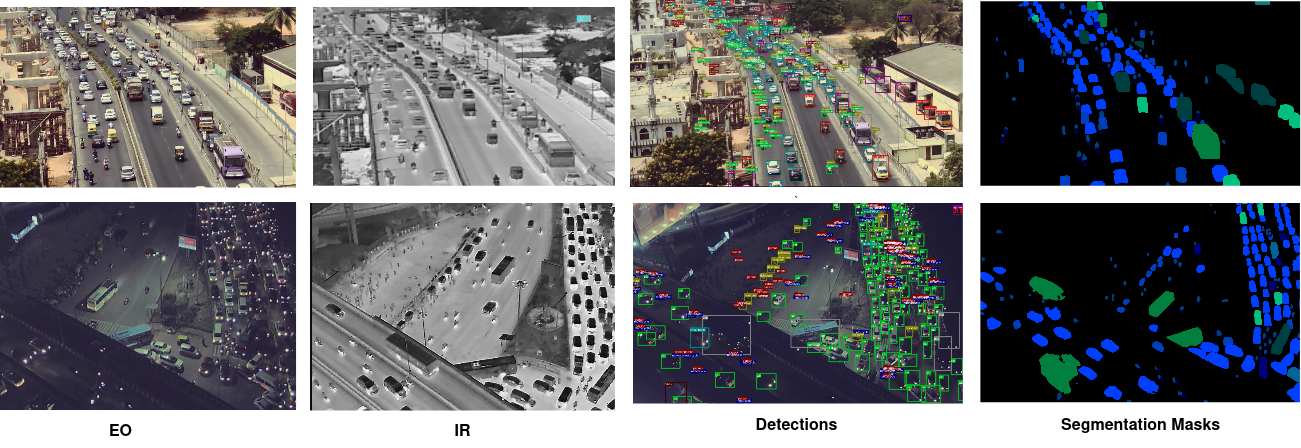}
    \caption{  Snapshots from the IndraEye dataset  showing different modalities EO, IR and complete semantic annotations for detection \& segmentation tasks taken from different slant angles}
    \label{fig:inra}
\end{figure*}

To address these challenges, we introduce the IndraEye dataset, a comprehensive open-source resource tailored for aerial perception through EO-IR drone imagery. This dataset fulfills the need for a versatile and multitask tool by providing an extensive collection of EO-IR images. It includes images of various categories, such as road vehicles and pedestrians, captured from different angles, backgrounds, and scales. Covering a broad range of environments—from bustling urban areas to expansive highways—it captures diverse scene characteristics, including population density, crowd dynamics, and environmental conditions. Collected using the DragonEye 2 drone mounted camera, which features an advanced gimbal and sensor array for both EO and IR imagery, each image is carefully annotated for multiple tasks. On average, EO images contain 35 annotated instances, with a particular focus on dense traffic scenarios. Examples of EO and IR images with different slant angles are presented in Figure \ref{fig:inra_dronw}. IndraEye is particularly valuable for tackling challenges related to domain adaptation, and open-vocabulary learning due to the notable differences between EO and IR images. Its support for multitask learning, including object detection and semantic segmentation, enables deep neural networks (DNNs) to develop more generalized representations. Consequently, IndraEye is set to be a pivotal resource for testing and enhancing model robustness across various modalities and tasks.
The salient contributions of our work are listed as follows:

\begin{itemize}
    \item We propose the IndraEye dataset, which features 7 diverse scenes with varying slant angles and height differences, resulting in significant scale variations of objects. The dataset includes 145,666 dense instances across 13 classes, including various road vehicles and people, in both EO and IR modalities (see Fig. \ref{fig:inra}).
    \item To the best of our knowledge, IndraEye is the first open-source EO-IR aerial object detection dataset. IndraEye is also the first open-source EO-IR aerial road object detection dataset procured in the country of India.
    \item We analyze the IndraEye dataset for object detection and segmentation and focusing on domain adaptation. Our study includes both vision and vision-language models, addressing challenges under fog and nighttime conditions, and evaluates VLMs' performance with slant angles and multi-modal data.
\end{itemize}


\begin{table*}[h]

\centering
\caption{Qualitative comparison of multiple aerial vehicle object detection datasets.} 
\label{table:Dataset comparison}
\begin{adjustbox}{max width=1.0\textwidth}
\begin{tabular}{l c c c c c c c}
\toprule
\textbf{Datasets} & \multicolumn{1}{l}{\textbf{Multi-sensory}} & \multicolumn{1}{l}{\textbf{Diverse Viewpoints}} & \multicolumn{1}{l}{\textbf{Diverse backgrounds}} & \multicolumn{1}{l}{\textbf{Diverse classes}} & \multicolumn{1}{l}{\textbf{Diverse illumination}} & \multicolumn{1}{l}{\textbf{Detection}} & \multicolumn{1}{l}{\textbf{Segmentation}}\\
\midrule
DOTA \cite{xia2018dota}             & \ding{55}                                          & \ding{55}                                                & \ding{55}                                                 & \checkmark                                             & \ding{55}                                    & \checkmark                 & \ding{55}          \\
HIT-UAV \cite{suo2023hit}         & \ding{55}                                           & \checkmark                                                & \checkmark                                                 & \ding{55}                                             & \ding{55}                             & \checkmark                   & \ding{55}               \\
VisDrone \cite{cao2021visdrone}        & \ding{55}                                           & \checkmark                                                & \checkmark                                                 & \checkmark                                             & \checkmark                        & \checkmark                   & \ding{55}                                       \\
UAVDT  \cite{du2018unmanned}           & \ding{55}                                           & \checkmark                                                & \checkmark                                                 & \checkmark                                             & \checkmark                        & \checkmark                   & \ding{55}                                       \\
Vedai \cite{razakarivony2016vehicle}            & \checkmark                                           & \ding{55}                                                & \checkmark                                                 & \checkmark                                             & \checkmark               & \checkmark                   & \ding{55}                                                \\
M3FD \cite{liu2022target}             & \checkmark                                           & \ding{55}                                                & \checkmark                                                 & \checkmark                                             & \checkmark                        & \checkmark                   & \ding{55}                                       \\
FLIR \cite{fliresh}            & \checkmark                                           & \ding{55}                                                & \checkmark                                                 & \checkmark                                             & \checkmark                               & \checkmark                   & \ding{55}                                \\
MSRS \cite{tang2022piafusion}          & \checkmark                                           & \ding{55}                                              & \checkmark                                                 & \checkmark                                             & \checkmark                                & \ding{55}                   & \checkmark                               \\
InfraParis \cite{franchi2024infraparis}  & \checkmark                                           & \ding{55}                                              & \checkmark                                                 & \checkmark                                             & \checkmark                                & \checkmark                   & \checkmark                               \\
{\textbf{IndraEye (Ours)}}   & \checkmark                                           & \checkmark                                                & \checkmark                                                 & \checkmark                                             & \checkmark                                & \checkmark                   & \checkmark                               \\
\bottomrule
\end{tabular}
\end{adjustbox}
\end{table*}
\section{RELATED WORK}
\textbf{Datasets for Aerial Object Detection} Recent research in object detection using aerial imagery has advanced significantly, driven by the availability of diverse datasets captured by Electro-Optical (EO) cameras. These datasets are typically categorized based on the viewing angle: (1) birds-eye or nadir-view datasets, such as VEDAI \cite{razakarivony2016vehicle} and DOTA \cite{xia2018dota}, and (2) slant-angle datasets, including VisDrone \cite{cao2021visdrone} and HIT-UAV \cite{suo2023hit}. Noteworthy contributions from datasets like UAVDT \cite{du2018unmanned} and \cite{cao2021visdrone} have highlighted the importance of UAV-based object detection and have driven progress in this field. These datasets are primarily used for applications such as traffic monitoring and aerial perception. Although they offer a range of images with varying viewpoints, backgrounds, and lighting conditions, object detection algorithms often struggle at night. While EO cameras provide high resolution and detailed texture information in well-lit conditions, their performance diminishes in low-light environments, where texture details are sparse, leading to reduced detection accuracy.\\
\textbf{Aerial Object Detection Using IR Imagery} To address low-illumination conditions from a dataset perspective, it is crucial to provide discriminative information that deep learning models can use to make reliable inferences. One effective approach is to use Infrared (IR) cameras, which capture thermal signatures to create digital images, thereby enabling the creation of datasets that help algorithms identify features in nighttime conditions. HIT-UAV \cite{suo2023hit} is an example of a dataset that utilizes UAV-based IR imagery for object detection. However, HIT-UAV has a limited number of classes and does not include RGB information, which could be valuable for detecting objects in daylight or well-lit scenarios.\\
\textbf{ Multi-modal datasets} Many EO-IR (multi-sensory) datasets, such as M3FD \cite{liu2022target}, FLIR \cite{fliresh}, and InfraParis \cite{franchi2024infraparis}, are mainly tailored for autonomous driving applications due to their low-altitude image capture and fixed viewpoints. This design limits their usefulness for UAV-based aerial perception tasks. In contrast, the Vedai \cite{razakarivony2016vehicle} dataset provides EO-IR paired images with diverse classes and lighting conditions suitable for aerial imagery. However, Vedai's major drawback for UAV-based object detection is its lack of coverage for multiple viewing angles, which are essential for effective detection from UAVs. In this work, we present the EO-IR (multi-sensor) dataset, IndraEye. This dataset includes a wide range of viewpoints, backgrounds, illumination conditions, and classes, reflecting various Indian traffic scenarios. IndraEye is designed to be a valuable resource for UAV-based object detection, supporting continuous aerial perception and traffic monitoring. To give a detailed comparison of existing Infrared datasets, we include an overview and comparison with our IndraEye dataset in Table \ref{table:Dataset comparison}.

\section{Aerial IndraEye Dataset}

\subsection{Acquisition process}
To create the IndraEye dataset, we used the DJI M600 Pro drone (Fig. 1), equipped with a DragonEye2 camera mounted on a gimbal to capture various slant-angle views by adjusting pitch and yaw. For low-altitude captures, the camera and gimbal setup were placed on a 3-meter elevated tripod. The DragonEye2 camera includes one Electro-Optical (EO) sensor and one uncooled Infrared (IR) sensor, with detailed specifications provided in Table 2. IndraEye was collected at different times of day—noon, evening, and night—to account for variations in illumination and backgrounds, thereby reducing inherent biases in neural networks (additional examples are available in the supplementary material). After image collection, the dataset was constructed by manually annotating target objects with bounding boxes in both EO and IR data. Videos from the EO-IR camera were recorded at seven locations in Bengaluru, including the Indian Institute of Science campus. Dataset at each location was collected for approximately 4 minutes, resulting in about 20,000 frames per scene. To improve object diversity and manage redundant content, every 35th frame was selected for inclusion in the dataset. 

\begin{table}[ht]
\centering
\caption{Sensor specification of the DragonEye 2 EO-IR camera.}
\label{sensor_specs}
\begin{tabular}{c c c l}
\toprule
\multicolumn{1}{l}{\textbf{Sensor}} & \multicolumn{1}{l}{\textbf{Resolution}} & \multicolumn{1}{l}{\textbf{Wavelength}} & \textbf{FoV} \\ 
\midrule
Visible camera                        & 1280x720                                 & 400-700\text{nm}                                & 60$^\circ$                      \\ 
Thermal Camera                        & 640x480                                  & 8-14$\mu$m                                     &  32$^\circ$                      \\ 
\bottomrule
\end{tabular}
\end{table}

\subsection{Camera calibration} 

Due to the hardware and software limitations of the camera, calibrating a system like DragonEye2 where both EO and IR cameras are integrated into a single module mounted on a drone proves to be challenging. This difficulty arises because the field of view (FoV) varies dynamically with the movement of the drone along with the camera, complicating the calibration of both cameras as mentioned in \cite{brenner2023rgb}. While stereo camera calibration is effective, it encounters a significant challenge from the different FOV's of the modalities (EO-IR) which can result in a parallax effect that varies at different depths. This phenomenon is due to the difference in viewing angles between the cameras, causing objects at various depths to appear at different positions in the image. Consequently, employing a single homography matrix, a transformation that maps points from one image to corresponding points in another is not very effective due to the variation in perspective. This misalignment becomes evident in the fused data when the vehicle with the camera is in motion \cite{guo2024damsdet}. Image alignment can lead to inaccuracies, particularly with long distance and small objects. As a result, manual co-registration is avoided. While the EO and IR images are captured with the same timestamp, they still exhibit slight misalignment. This limitation suggests that multi-modal applications, which do not require co-registered images, such as domain adaptation techniques \cite{do2024d3t}, \cite{gan2023unsupervised} can be explored.

\subsection{Statistics of the dataset} 
IndraEye comprises 5,612 images captured at various locations throughout Bengaluru. The dataset includes multiple viewing angles, altitudes, backgrounds, and times of day. The EO images are divided into 2,336 samples (2,026 for training, 60 for validation, and 250 for testing), while the IR images are divided into 3,276 samples (2,973 for training, 58 for validation, and 245 for testing). The dataset also includes day and night splits if required. The goal was to create a diverse and comprehensive dataset with both EO and IR images, suitable for various conditions and contexts.
IndraEye features 13 classes commonly found in the Southern Asian subcontinent: backhoe loader, bicycle, bus, car, cargo truck, cargo trike (a medium-sized three-wheeled cargo vehicle), ignore, motorcycle, person, rickshaw (a small three-wheeled passenger vehicle), small truck, truck, tractor, and van. Unlike existing datasets, which predominantly focus on Western countries with minimal use of three-wheeled vehicles and less dense traffic, IndraEye captures the unique traffic patterns of the region. Detailed class distribution for the IndraEye training set is shown in Table \ref{table3}. During the image capture process, some images were deemed unusable. To enhance object diversity and manage redundancy, every 35th frame from the 20,000 frames captured per scene was selected for inclusion.

\begin{table*}[h]
\centering
\caption{IndraEye dataset description: with altitude of imaged scene, dynamic angle ranges, scene-wise instances, and scale-variability of each scene.}
\label{dataset_desc}
\begin{adjustbox}{max width=1.0\textwidth}
\begin{tabular}{c c c c c c c c}
\toprule
\textbf{Scene} & \textbf{\begin{tabular}[c]{@{}c@{}}Altitude\\ (metres)\end{tabular}} & \textbf{\begin{tabular}[c]{@{}c@{}}Dynamic Angle \\ range (degrees)\end{tabular}} & \textbf{\begin{tabular}[c]{@{}c@{}}EO Instances\\ daytime\end{tabular}} & \textbf{\begin{tabular}[c]{@{}c@{}}EO Instances\\ nightime\end{tabular}} & \textbf{\begin{tabular}[c]{@{}c@{}}IR Instances\\ daytime\end{tabular}} & \textbf{\begin{tabular}[c]{@{}c@{}}IR Instances\\ nightime\end{tabular}} & \textbf{Scale-variablity} \\
\midrule
A   & 30                                                                   & 10-25                                                                   & 1982                                                                    & -                                                                        & 9247                                                                    & -                                                                        & Mid                       \\ 
B   & 30                                                                   & 10-25                                                                   & 14149                                                                   & -                                                                        & 8215                                                                    & -                                                                        & Mid                       \\ 
C         & 60                                                                   & 5-50                                                                    & 37752                                                                   & 5312                                                                     & 7559                                                                    & 1285                                                                     & High                      \\ 
D         & 12                                                                   & 20-40                                                                   & 5369                                                                    & -                                                                        & -                                                                       & -                                                                        & Low                       \\ 
E        & 12                                                                   & 20-40                                                                   & 5394                                                                    & 3774                                                                     & -                                                                       & 6124                                                                     & Low                       \\ 
F         & 12                                                                   & 10-30                                                                   & 4728                                                                    & -                                                                        & 8759                                                                    & -                                                                        & Mid                       \\ 
G            & 7                                                                    & 20-40                                                                   & 2234                                                                    & -                                                                        & 1936                                                                    & 2024                                                                     & Low                       \\ 
H            & 7                                                                    & 10-30                                                                   & 3033                                                                    & 2971                                                                     & 3731                                                                    & 7413                                                                     & High                      \\  
\bottomrule
\end{tabular}
\end{adjustbox}
\label{table3}
\end{table*}

\subsection{Class labels} 

For the IndraEye dataset, we provide ground-truth labels for both object detection and pixel-level annotations across all classes using a two-stage approach.

\begin{enumerate}
  \item  \textbf{Zero-Shot Annotations with Human in the Loop:}
    In this initial step, we generate zero-shot annotations by utilizing pre-trained models. These models are used to produce preliminary annotations with human oversight, ensuring that the annotations align closely with the true object locations and characteristics.
    \item \textbf{Manual Verification and Refinement:}
    After generating the initial annotations, each label undergoes thorough manual verification. This step involves a detailed review of the annotations to correct any discrepancies and refine the labels for higher accuracy.
\end{enumerate}

This structured approach ensures that both object detection and pixel-level annotations are meticulously crafted and reliable. For the object detection task, EO and IR images are manually annotated with bounding boxes using the X-AnyLabeling tool \cite{Wang_X-AnyLabeling}. This tool facilitates efficient annotation by allowing users to load pre-trained models, such as those from the YOLO family of networks. After multiple iterations of correction and evaluation, highly precise annotations have been generated. For semantic segmentation, drawing on the success of SAM \cite{kirillov2023segment} in generating segmentation masks with zero-shot performance, and considering that IndraEye is an aerial imagery dataset, we employ SAMRS \cite{wang2024samrs}. SAMRS builds on SAM \cite{kirillov2023segment} and integrates existing remote sensing datasets to create an efficient pipeline for generating masks for large-scale remote sensing segmentation datasets. This approach has streamlined our process for producing pixel-level annotations for IndraEye. The annotations for both tasks are carefully designed to match the established class schema.

\subsection{Ethics and policy} The proposed dataset undergoes a thorough manual review to ensure privacy protection. Faces of individuals and vehicle license plates that are clearly visible in the images are blurred to maintain confidentiality and safeguard personal rights. This process ensures that individuals' privacy is upheld while preserving the dataset's utility.

\section{EXPERIMENTAL ANALYSIS}

This section describes experiments conducted to assess the advantages of using EO-IR sensors for drone-based object detection, segmentation. We explore various settings, including domain adaptation and generalization. The experiments are performed on an Nvidia A6000 GPU with 48GB of memory.


\subsection{Object Detection}

We evaluate the IndraEye dataset on state-of-the-art aerial object detection models to showcase the performance of these models when there is a mixture of scale-variability, diverse illumination and background conditions. The in-domain performance of these models falls significantly short of the standards required for deployment in real-world, safety-critical applications. This highlights the necessity for such datasets and the development of model architectures that are both generalizable and adaptable to a wide range of scenarios. 

Table \ref{fig:detection-aerial} evaluates the performance of Yolov8x \cite{Jocher_Ultralytics_YOLO_2023} on IndraEye on five different configurations. The  model is trained at Yolov8x’s default hyper-parameter settings with epoch set to 100. These five configurations showcases the limitations of each sensor modality and promotes the use of both the sensory modalities for drone-based aerial perception. The four configurations include: Firstly, training and testing on the EO modality with both day and nighttime images to mimic real-world situations for sensory systems with only a RGB/EO sensor. The first configuration shows the pitfall of the EO imagery when tested on low-lit condition.
The second configuration evaluates the effectiveness of using only EO sensors for aerial perception purposes. The third configuration is completely based on the capability of the model learning from the IR imagery. Precisely, in this setting IR modality with both day and nighttime images are used to train the DNN. As can be seen from the table \ref{fig:detection-aerial}, comparison from the first three test configurations, it is clear that the IR modality is better suited for low-illumination conditions like nighttime. The fourth and fifth configurations are the counterparts for the second and third configurations, wherein, the evaluation is conducted under well-illuminated (daytime) conditions. The results clearly suggest that the EO sensor modality captures rich texture information is better suited in such conditions as compared to the IR modality which is more suited towards lowly-lit conditions due to its functionality of capturing thermal emmissivity from the environment to form an image. 

Table \ref{fig:detection-dota} assesses the performance of state-of-the-art detection models on nadir-view aerial imagery. While these models demonstrate high performance on other datasets, their effectiveness on the IndraEye dataset is relatively limited. Specifically for EO images, the challenges of densely packed objects, varying slant angles, and diverse illumination conditions contribute to the restricted performance on the IndraEye dataset.

\begin{table}[h]
\centering
\caption{Performance of object detection algorithms on the proposed IndraEye dataset}
\label{table:Dataset comparison}
\begin{adjustbox}{max width=0.5\textwidth}
\begin{tabular}{@{}c c c c c c@{}}
\toprule
\textbf{Models} & \textbf{Train on EO} & \textbf{Test on EO} & \textbf{Train on IR} & \textbf{Test on IR} & \textbf{mAP50} \\ 
\midrule
FasterRCNN \cite{NIPS2015_14bfa6bb}     & \checkmark                   & \checkmark                  & \ding{55}                   & \ding{55}                  & 47.6 \\ 

ReDet \cite{han2021redet}          & \checkmark                   & \checkmark                  & \ding{55}                   & \ding{55}                  & 43.9 \\ 

ORCNN \cite{xie2021oriented}          & \checkmark                   & \checkmark                  & \ding{55}                   & \ding{55}                  & 56.3 \\ 

Faster RCNN \cite{NIPS2015_14bfa6bb}    & \ding{55}                   & \ding{55}                  & \checkmark                   & \checkmark                  & 57.3 \\ 

ReDet \cite{han2021redet}          & \ding{55}                   & \ding{55}                  & \checkmark                   & \checkmark                  & 43.8 \\ 

ORCNN \cite{xie2021oriented}          & \ding{55}                   & \ding{55}                  & \checkmark                   & \checkmark                  & 65.6 \\ 
\bottomrule
\end{tabular}
\end{adjustbox}
\label{fig:detection-dota}
\end{table}

\subsection{Semantic Segmentation}

As shown in table \ref{fig:segmentation}, we evaluate the performance of semantic segmentation tasks using models such as UNet \cite{ronneberger2015u}, DeepLabV3+ \cite{chen2017rethinking}, DeepMao \cite{sikdar2023deepmao}, and Mask2Former \cite{sikdar2023deepmao} with various backbones. These models are trained on EO imagery and then applied to both EO and IR modalities from the IndraEye dataset. For the IR modality, the results are reported in a zero-shot setup, where the models are trained exclusively on EO images and then tested on IR images. This configuration showcases the models' ability to generalize effectively across domain shifts. We maintain a batch size of 8 across all settings and train each model for 15 epochs. UNet delivers the highest mIoU for EO inference, while DeepMao surpasses the other models in IR inference. This is noteworthy because DeepMao was initially trained on aerial imagery for detecting small building footprints, suggesting that the model has effectively adapted to the IndraEye dataset due to shared characteristics like occlusions. Although there is a performance drop in IR inference due to domain shifts, the results indicate that there is potential for further adaptation to better handle this spectral shift.

\begin{table}[ht]
\centering
\caption{Performance of segmentation models on the IndraEye dataset when trained on the EO train split.}
\label{Seg table}
\begin{tabular}{c c c}
\toprule
\textbf{Models}    & \textbf{mIoU (EO)} & \textbf{mIoU (IR)} \\ \hline
UNet (ResNet-50)\cite{ronneberger2015u}     & 80.75     & 41.56     \\ 
FPN (ResNet-50)\cite{lin2017feature}      & 79.18     & 40.59     \\ 
DeepMao (EfficientNet-B3)\cite{sikdar2023deepmao} & 80.70     & 47.12     \\ 
DeeplabV3 (ResNet-50)\cite{chen2017rethinking}   & 79.23     & 41.45     \\ 
Mask2Former (Swin-B)\cite{sikdar2023deepmao}   & 65.0     & 22.13    \\ 
\bottomrule
\end{tabular}
\label{fig:segmentation}
\end{table}

\begin{table}[ht]
\centering
\caption{Unsupervised Domain Adaptation from EO modality to IR modality}
\label{UDA table}
\begin{tabular}{@{}lcc@{}}
\toprule
\multicolumn{1}{c}{\textbf{Method}}                                     & \multicolumn{2}{c}{\textbf{mAP50}} \\
\multicolumn{1}{l}{}                     & \multicolumn{1}{c}{\textbf{FLIR Dataset}} & \multicolumn{1}{c}{\textbf{IndraEye Dataset}} \\ \midrule
Source (Faster-RCNN) \cite{NIPS2015_14bfa6bb}                     & 23.2                             & 21.4                             \\ 
AT \cite{li2022cross}                                         & 26.5                             & 23.2                             \\ 
CMT \cite{cao2023contrastive}                                      & 28.4                             & 28.0                             \\ 
Oracle (Faster-RCNN) \cite{NIPS2015_14bfa6bb}                                   & 35.0                             & 58.1                             \\ \bottomrule
\end{tabular}
\end{table}
\subsection{Domain adaptation and generalization}
To address the gap between the two modalities, which are not pixel-to-pixel coregistered, domain adaptation techniques are explored. In particular, we conduct experiments with state-of-the-art models that utilize common knowledge shared between the modalities, leading to improved performance on downstream tasks in the target domain (IR). By leveraging this shared knowledge, the goal is to learn domain-invariant features that enable the model to generalize effectively across different domains.

To evaluate the performance of UDA methods, we use AT \cite{li2022cross} and CMT \cite{cao2023contrastive} with VGG-16 \cite{simonyan2014very} backbone. We run the adaptation models for 50k iterations with batch size being set to 32.
Table \ref{UDA table} compares the performance of SOTA UDA methods on the FLIR and IndraEye datasets adapting from EO to IR modality. The first three settings involve inference on IR data, while the oracle setting is a train-on-IR, test-on-IR scenario. It is evident that all the UDA methods perform better on the FLIR dataset than the proposed IndraEye dataset. Additionally, for the FLIR dataset, the performance gap between the Oracle setting and the best-performing UDA method is smaller compared to the IndraEye dataset, indicating that domain adaptation on IndraEye is a more challenging task. This is primarily due to the occlusions and scale-variation in the feature-space of the high-look angle imagery which is not accounted for in the FLIR dataset.

\begin{table}[h]
\centering
\caption{Generalization of object detection model trained on different aerial EO imagery datasets}
\label{DG}
\begin{tabular}{@{}lll@{}}
\toprule
\textbf{Train set} & \textbf{Test set} & \textbf{mAP50} \\ 
\midrule
VisDrone           & VisDrone          & 53.3           \\ 
IndraEye           & VisDrone          & 32.5           \\ 
IndraEye           & IndraEye          & 83.135         \\ 
VisDrone           & IndraEye          & 52.085         \\
\bottomrule
\end{tabular}
\end{table}

Due to its diverse conditions across multiple sensor modalities and high-quality ground-truth data, the IndraEye dataset serves as an ideal testbed for model adaptation and generalization experiments, particularly in the domain of aerial perception. To benchmark other aerial perception datasets against IndraEye in the EO domain, we trained YOLOv8x on the VisDrone dataset, which includes significantly more training image instances than IndraEye. For a fair comparison, we limited the analysis to the common classes shared between the two datasets (5 classes). Despite the larger number of training instances in VisDrone (6471 number of training samples in Visdrone compared to 2026 number of training samples in IndraEye), as shown in Table \ref{DG}, the difference between the oracle setting (VisDrone to VisDrone/IndraEye to IndraEye) and the cross-domain gap clearly indicates that the IndraEye dataset seems to help the model generalize better than the VisDrone dataset. We attribute this improvement to the greater diversity and higher quality of annotations in IndraEye. Nevertheless, the significant performance drop of all state-of-the-art unsupervised domain adaptation (UDA) methods under domain shifts underscores the ongoing need for advancements in this field.

\section{CONCLUSION}
In this paper, we introduce IndraEye, the first multi-sensor aerial perception dataset designed for less illuminated conditions in the Indian subcontinent. Covering a wide range of environments—from bustling urban areas to expansive highways, the dataset captures diverse scene characteristics, such as population density, crowd dynamics, and environmental conditions. IndraEye includes EO images with 2,336 samples and IR images with 3,276 samples, featuring multiple viewing angles, altitudes, backgrounds, and times of day. The dataset is valuable for training and provides a rigorous evaluation for models in varied visual conditions and densely cluttered environments. Our experimental analysis positions IndraEye as a challenging benchmark for both uni- and multimodal aerial perception tasks, including domain adaptation and zero-shot learning. The dataset opens up several research avenues, including multimodal learning, domain adaptation for object detection and segmentation, exploration of sensor-specific strengths and weaknesses, and model generalization or adaptation across different conditions and modalities.

\begin{table}[]
\caption{Table illustrating how the features of the two modalities complement each other, utilizing a day-night split within the train-test split of the IndraEye dataset.}
\label{EO-D+N}
\centering
\begin{tabular}{c c c}
\hline
\textbf{\begin{tabular}[c]{@{}c@{}}Train\\ Setting\end{tabular}} & \textbf{\begin{tabular}[c]{@{}c@{}}Test\\ Configuration\end{tabular}} & \textbf{mAP50} \\ \hline
EO-Day                                                     & EO-Night                                                                  & 30.0            \\ 
EO-Day + EO-Night                                                      & EO-Night                                                                  & 52.0            \\ 
IR-Day + IR-Night                                                      & IR-Night                                                                  & 73.6           \\ 
EO-Day  + EO-Night                                                      & EO-Day                                                                  & 90.7           \\ 
IR-Day + IR-Night                                                      & IR-Day                                                                  & 77.0            \\ 
\bottomrule
\end{tabular}
\label{fig:detection-aerial}
\end{table}

\addtolength{\textheight}{-12cm}   



\bibliographystyle{plain}
\bibliography{iros24}

\begin{thebibliography}{10}

\bibitem{akkaya2021self}
Ibrahim~Batuhan Akkaya, Fazil Altinel, and Ugur Halici.
\newblock Self-training guided adversarial domain adaptation for thermal imagery.
\newblock In {\em Proceedings of the IEEE/CVF conference on computer vision and pattern recognition}, pages 4322--4331, 2021.

\bibitem{bozcan2020air}
Ilker Bozcan and Erdal Kayacan.
\newblock Au-air: A multi-modal unmanned aerial vehicle dataset for low altitude traffic surveillance.
\newblock In {\em 2020 IEEE International Conference on Robotics and Automation (ICRA)}, pages 8504--8510. IEEE, 2020.

\bibitem{bravo2019use}
Raissa Zurli~Bittencourt Bravo, Adriana Leiras, and Fernando~Luiz Cyrino~Oliveira.
\newblock The use of uavs in humanitarian relief: An application of pomdp-based methodology for finding victims.
\newblock {\em Production and Operations Management}, 28(2):421--440, 2019.

\bibitem{brenner2023rgb}
Martin Brenner, Napoleon~H Reyes, Teo Susnjak, and Andre~LC Barczak.
\newblock Rgb-d and thermal sensor fusion: a systematic literature review.
\newblock {\em IEEE Access}, 2023.

\bibitem{cao2023contrastive}
Shengcao Cao, Dhiraj Joshi, Liang-Yan Gui, and Yu-Xiong Wang.
\newblock Contrastive mean teacher for domain adaptive object detectors.
\newblock In {\em Proceedings of the IEEE/CVF Conference on Computer Vision and Pattern Recognition}, pages 23839--23848, 2023.

\bibitem{cao2021visdrone}
Yaru Cao, Zhijian He, Lujia Wang, Wenguan Wang, Yixuan Yuan, Dingwen Zhang, Jinglin Zhang, Pengfei Zhu, Luc Van~Gool, Junwei Han, et~al.
\newblock Visdrone-det2021: The vision meets drone object detection challenge results.
\newblock In {\em Proceedings of the IEEE/CVF International conference on computer vision}, pages 2847--2854, 2021.

\bibitem{chen2017rethinking}
Liang-Chieh Chen, George Papandreou, Florian Schroff, and Hartwig Adam.
\newblock Rethinking atrous convolution for semantic image segmentation. arxiv.
\newblock {\em arXiv preprint arXiv:1706.05587}, 5, 2017.

\bibitem{chen2022neighbortrack}
Yu-Hsi Chen, Chien-Yao Wang, Cheng-Yun Yang, Hung-Shuo Chang, Youn-Long Lin, Yung-Yu Chuang, and Hong-Yuan~Mark Liao.
\newblock Neighbortrack: Improving single object tracking by bipartite matching with neighbor tracklets.
\newblock {\em arXiv preprint arXiv:2211.06663}, 2022.

\bibitem{cho2024cat}
Seokju Cho, Heeseong Shin, Sunghwan Hong, Anurag Arnab, Paul~Hongsuck Seo, and Seungryong Kim.
\newblock Cat-seg: Cost aggregation for open-vocabulary semantic segmentation.
\newblock In {\em Proceedings of the IEEE/CVF Conference on Computer Vision and Pattern Recognition}, pages 4113--4123, 2024.

\bibitem{do2024d3t}
Dinh~Phat Do, Taehoon Kim, Jaemin Na, Jiwon Kim, Keonho Lee, Kyunghwan Cho, and Wonjun Hwang.
\newblock D3t: Distinctive dual-domain teacher zigzagging across rgb-thermal gap for domain-adaptive object detection.
\newblock In {\em Proceedings of the IEEE/CVF Conference on Computer Vision and Pattern Recognition}, pages 23313--23322, 2024.

\bibitem{du2018unmanned}
Dawei Du, Yuankai Qi, Hongyang Yu, Yifan Yang, Kaiwen Duan, Guorong Li, Weigang Zhang, Qingming Huang, and Qi~Tian.
\newblock The unmanned aerial vehicle benchmark: Object detection and tracking.
\newblock In {\em Proceedings of the European conference on computer vision (ECCV)}, pages 370--386, 2018.

\bibitem{franchi2024infraparis}
Gianni Franchi, Marwane Hariat, Xuanlong Yu, Nacim Belkhir, Antoine Manzanera, and David Filliat.
\newblock Infraparis: A multi-modal and multi-task autonomous driving dataset.
\newblock In {\em Proceedings of the IEEE/CVF Winter Conference on Applications of Computer Vision}, pages 2973--2983, 2024.

\bibitem{gan2023unsupervised}
Lu~Gan, Connor Lee, and Soon-Jo Chung.
\newblock Unsupervised rgb-to-thermal domain adaptation via multi-domain attention network.
\newblock In {\em 2023 IEEE International Conference on Robotics and Automation (ICRA)}, pages 6014--6020. IEEE, 2023.

\bibitem{guo2024damsdet}
Junjie Guo, Chenqiang Gao, Fangcen Liu, Deyu Meng, and Xinbo Gao.
\newblock Damsdet: Dynamic adaptive multispectral detection transformer with competitive query selection and adaptive feature fusion.
\newblock {\em arXiv e-prints}, pages arXiv--2403, 2024.

\bibitem{han2021redet}
Jiaming Han, Jian Ding, Nan Xue, and Gui-Song Xia.
\newblock Redet: A rotation-equivariant detector for aerial object detection.
\newblock In {\em Proceedings of the IEEE/CVF conference on computer vision and pattern recognition}, pages 2786--2795, 2021.

\bibitem{he2024prompting}
Qibin He.
\newblock Prompting multi-modal image segmentation with semantic grouping.
\newblock In {\em Proceedings of the AAAI Conference on Artificial Intelligence}, volume~38, pages 2094--2102, 2024.

\bibitem{heintz2007images}
Fredrik Heintz, Piotr Rudol, and Patrick Doherty.
\newblock From images to traffic behavior-a uav tracking and monitoring application.
\newblock In {\em 2007 10th International Conference on Information Fusion}, pages 1--8. IEEE, 2007.

\bibitem{hsieh2017drone}
Meng-Ru Hsieh, Yen-Liang Lin, and Winston~H Hsu.
\newblock Drone-based object counting by spatially regularized regional proposal network.
\newblock In {\em Proceedings of the IEEE international conference on computer vision}, pages 4145--4153, 2017.

\bibitem{Jocher_Ultralytics_YOLO_2023}
Glenn Jocher, Ayush Chaurasia, and Jing Qiu.
\newblock {Ultralytics YOLO}, January 2023.

\bibitem{kanistras2013survey}
Konstantinos Kanistras, Goncalo Martins, Matthew~J Rutherford, and Kimon~P Valavanis.
\newblock A survey of unmanned aerial vehicles (uavs) for traffic monitoring.
\newblock In {\em 2013 international conference on unmanned aircraft systems (ICUAS)}, pages 221--234. IEEE, 2013.

\bibitem{kirillov2023segment}
Alexander Kirillov, Eric Mintun, Nikhila Ravi, Hanzi Mao, Chloe Rolland, Laura Gustafson, Tete Xiao, Spencer Whitehead, Alexander~C Berg, Wan-Yen Lo, et~al.
\newblock Segment anything.
\newblock In {\em Proceedings of the IEEE/CVF International Conference on Computer Vision}, pages 4015--4026, 2023.

\bibitem{li2022cross}
Yu-Jhe Li, Xiaoliang Dai, Chih-Yao Ma, Yen-Cheng Liu, Kan Chen, Bichen Wu, Zijian He, Kris Kitani, and Peter Vajda.
\newblock Cross-domain adaptive teacher for object detection.
\newblock In {\em Proceedings of the IEEE/CVF Conference on Computer Vision and Pattern Recognition}, pages 7581--7590, 2022.

\bibitem{lin2017feature}
Tsung-Yi Lin, Piotr Doll{\'a}r, Ross Girshick, Kaiming He, Bharath Hariharan, and Serge Belongie.
\newblock Feature pyramid networks for object detection.
\newblock In {\em Proceedings of the IEEE conference on computer vision and pattern recognition}, pages 2117--2125, 2017.

\bibitem{liu2022target}
Jinyuan Liu, Xin Fan, Zhanbo Huang, Guanyao Wu, Risheng Liu, Wei Zhong, and Zhongxuan Luo.
\newblock Target-aware dual adversarial learning and a multi-scenario multi-modality benchmark to fuse infrared and visible for object detection.
\newblock In {\em Proceedings of the IEEE/CVF conference on computer vision and pattern recognition}, pages 5802--5811, 2022.

\bibitem{paramanandham2018infrared}
Nirmala Paramanandham and Kishore Rajendiran.
\newblock Infrared and visible image fusion using discrete cosine transform and swarm intelligence for surveillance applications.
\newblock {\em Infrared Physics \& Technology}, 88:13--22, 2018.

\bibitem{razakarivony2016vehicle}
Sebastien Razakarivony and Frederic Jurie.
\newblock Vehicle detection in aerial imagery: A small target detection benchmark.
\newblock {\em Journal of Visual Communication and Image Representation}, 34:187--203, 2016.

\bibitem{NIPS2015_14bfa6bb}
Shaoqing Ren, Kaiming He, Ross Girshick, and Jian Sun.
\newblock Faster r-cnn: Towards real-time object detection with region proposal networks.
\newblock In C.~Cortes, N.~Lawrence, D.~Lee, M.~Sugiyama, and R.~Garnett, editors, {\em Advances in Neural Information Processing Systems}, volume~28. Curran Associates, Inc., 2015.

\bibitem{ronneberger2015u}
Olaf Ronneberger, Philipp Fischer, and Thomas Brox.
\newblock U-net: Convolutional networks for biomedical image segmentation.
\newblock In {\em Medical image computing and computer-assisted intervention--MICCAI 2015: 18th international conference, Munich, Germany, October 5-9, 2015, proceedings, part III 18}, pages 234--241. Springer, 2015.

\bibitem{saadiyean2024learning}
Qiranul Saadiyean, SP~Samprithi, and Suresh Sundaram.
\newblock Learning multi-scale context mask-rcnn network for slant angled aerial imagery in instance segmentation in a sim2real setup.
\newblock In {\em 2024 IEEE International Conference on Robotics and Automation (ICRA)}, pages 13573--13580. IEEE, 2024.

\bibitem{samad2013potential}
Abd~Manan Samad, Nazrin Kamarulzaman, Muhammad~Asyraf Hamdani, Thuaibatul~Aslamiah Mastor, and Khairil~Afendy Hashim.
\newblock The potential of unmanned aerial vehicle (uav) for civilian and mapping application.
\newblock In {\em 2013 IEEE 3rd International Conference on System Engineering and Technology}, pages 313--318. IEEE, 2013.

\bibitem{sikdar2024skd}
Aniruddh Sikdar, Jayant Teotia, and Suresh Sundaram.
\newblock Skd-net: Spectral-based knowledge distillation in low-light thermal imagery for robotic perception.
\newblock In {\em 2024 IEEE International Conference on Robotics and Automation (ICRA)}, pages 9041--9047. IEEE, 2024.

\bibitem{sikdar2023deepmao}
Aniruddh Sikdar, Sumanth Udupa, Prajwal Gurunath, and Suresh Sundaram.
\newblock Deepmao: Deep multi-scale aware overcomplete network for building segmentation in satellite imagery.
\newblock In {\em Proceedings of the IEEE/CVF Conference on Computer Vision and Pattern Recognition}, pages 487--496, 2023.

\bibitem{simonyan2014very}
Karen Simonyan and Andrew Zisserman.
\newblock Very deep convolutional networks for large-scale image recognition.
\newblock In {\em Proceedings of the IEEE Conference on Computer Vision and Pattern Recognition (CVPR)}, 2015.

\bibitem{fliresh}
Open source FLIR dataset~available online.
\newblock Free teledyne flir thermal dataset for algorithm training.
\newblock {\em https://www.flir.com/oem/adas/adas-dataset-form/}.

\bibitem{suo2023hit}
Jiashun Suo, Tianyi Wang, Xingzhou Zhang, Haiyang Chen, Wei Zhou, and Weisong Shi.
\newblock Hit-uav: A high-altitude infrared thermal dataset for unmanned aerial vehicle-based object detection.
\newblock {\em Scientific Data}, 10(1):227, 2023.

\bibitem{tang2022piafusion}
Linfeng Tang, Jiteng Yuan, Hao Zhang, Xingyu Jiang, and Jiayi Ma.
\newblock Piafusion: A progressive infrared and visible image fusion network based on illumination aware.
\newblock {\em Information Fusion}, 83:79--92, 2022.

\bibitem{wang2024samrs}
Di~Wang, Jing Zhang, Bo~Du, Minqiang Xu, Lin Liu, Dacheng Tao, and Liangpei Zhang.
\newblock Samrs: Scaling-up remote sensing segmentation dataset with segment anything model.
\newblock {\em Advances in Neural Information Processing Systems}, 36, 2024.

\bibitem{Wang_X-AnyLabeling}
Wei Wang.
\newblock {X-AnyLabeling}.

\bibitem{weir2019spacenet}
Nicholas Weir, David Lindenbaum, Alexei Bastidas, Adam~Van Etten, Sean McPherson, Jacob Shermeyer, Varun Kumar, and Hanlin Tang.
\newblock Spacenet mvoi: A multi-view overhead imagery dataset.
\newblock In {\em Proceedings of the ieee/cvf international conference on computer vision}, pages 992--1001, 2019.

\bibitem{xia2018dota}
Gui-Song Xia, Xiang Bai, Jian Ding, Zhen Zhu, Serge Belongie, Jiebo Luo, Mihai Datcu, Marcello Pelillo, and Liangpei Zhang.
\newblock Dota: A large-scale dataset for object detection in aerial images.
\newblock In {\em Proceedings of the IEEE conference on computer vision and pattern recognition}, pages 3974--3983, 2018.

\bibitem{xie2021oriented}
Xingxing Xie, Gong Cheng, Jiabao Wang, Xiwen Yao, and Junwei Han.
\newblock Oriented r-cnn for object detection.
\newblock In {\em Proceedings of the IEEE/CVF international conference on computer vision}, pages 3520--3529, 2021.

\bibitem{zhao2023cddfuse}
Zixiang Zhao, Haowen Bai, Jiangshe Zhang, Yulun Zhang, Shuang Xu, Zudi Lin, Radu Timofte, and Luc Van~Gool.
\newblock Cddfuse: Correlation-driven dual-branch feature decomposition for multi-modality image fusion.
\newblock In {\em Proceedings of the IEEE/CVF conference on computer vision and pattern recognition}, pages 5906--5916, 2023.

\end{thebibliography}
\end{document}